\documentclass[]{OAGM}
\OAGMarXiv{1404.3538}

\usepackage{graphics}
\usepackage{graphicx}
\usepackage{amsmath}
\usepackage{caption}
\usepackage{subcaption}
\usepackage{tikz}
\usepackage{booktabs}
\usepackage{color}

\usepackage{multirow}
\usepackage{tabulary}
\usepackage{colortbl} 
\usepackage{xparse}

\NewDocumentCommand{\rot}{O{90} O{1em} m}{\makebox[#2][l]{\rotatebox{#1}{#3}}}%

\definecolor{floor}{RGB}{228,26,28}
\definecolor{wall}{RGB}{55,126,184}
\definecolor{ceiling}{RGB}{255,127,0}
\definecolor{table}{RGB}{77,175,74}
\definecolor{chair}{RGB}{247,129,191}
\definecolor{cabinet}{RGB}{255,255,51}
\definecolor{object}{RGB}{152,78,163}
\definecolor{unknown}{RGB}{166,86,40}

\DeclareMathOperator*{\argmin}{arg\,min}

\title{Find my mug: Efficient object search with a mobile robot using semantic segmentation}
\newcommand{\shorttitle}{Efficient object search with a mobile robot using semantic segmentation}

\author{Daniel Wolf, Markus Bajones, Johann Prankl and Markus Vincze \\
Vision4Robotics Group, Automation \& Control Institute\\Vienna University of Technology}

\begin{document}
\maketitle

\setlength{\belowdisplayskip}{5pt} \setlength{\belowdisplayshortskip}{0pt}
\setlength{\abovedisplayskip}{5pt} \setlength{\abovedisplayshortskip}{0pt}

\begin{abstract}
In this paper, we propose an efficient semantic segmentation framework for indoor scenes, tailored to the application on a mobile robot. Semantic segmentation can help robots to gain a reasonable understanding of their environment, but to reach this goal, the algorithms not only need to be accurate, but also fast and robust. Therefore, we developed an optimized 3D point cloud processing framework based on a Randomized Decision Forest, achieving competitive results at sufficiently high frame rates. We evaluate the capabilities of our method on the popular NYU depth dataset and our own data and demonstrate its feasibility by deploying it on a mobile service robot, for which we could optimize an object search procedure using our results.

\end{abstract}

\vspace{-1.7mm}
\section{Introduction}
\vspace{-1.7mm}
It is the ultimate goal in the field of service robotics that mobile robots autonomously navigate and get along in human-made environments. A crucial step on the way to achieve this ambitious goal is that robots are able to recognize and interpret their surroundings.
Imagine a simple scenario where you ask your service robot to look for your mug. So far, in most applications the only knowledge the machine has about its environment is a simple 2D map encoding occupied and free space. That is, the only way to solve this task would be to execute a time-consuming brute-force object detection everywhere in the map. Would it not be much more intelligent if it first looked at the most probable locations for the mug to be, e.g.\ on tables or in the cupboard?
An important cornerstone to develop more intelligent behavior like this is semantic segmentation, which enables the robot to infer more meaningful information from its perceived environment.
\begin{figure}[bp]
  \centering
   \includegraphics[width=0.24\textwidth]{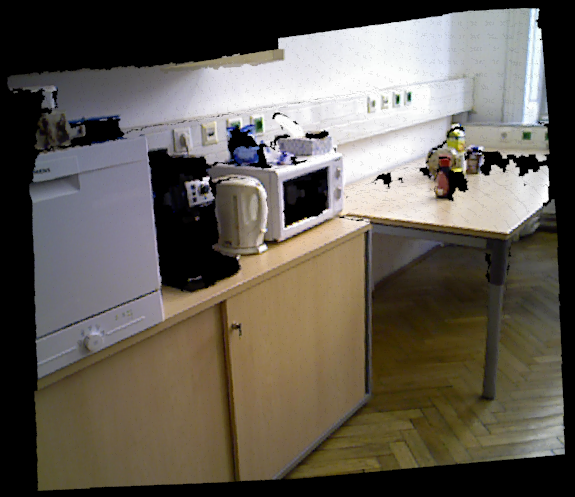} 
    \includegraphics[width=0.24\textwidth]{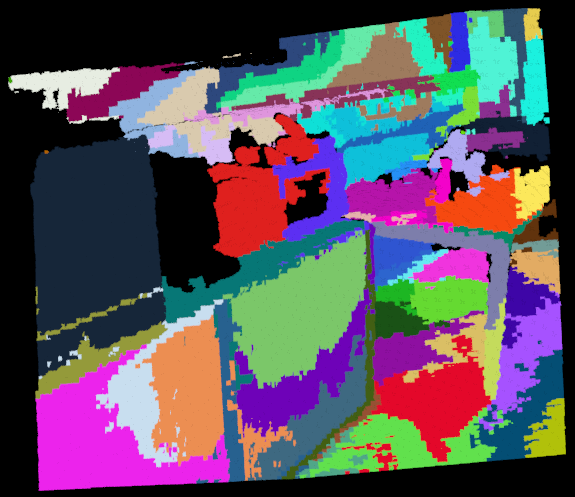} 
    \includegraphics[width=0.24\textwidth]{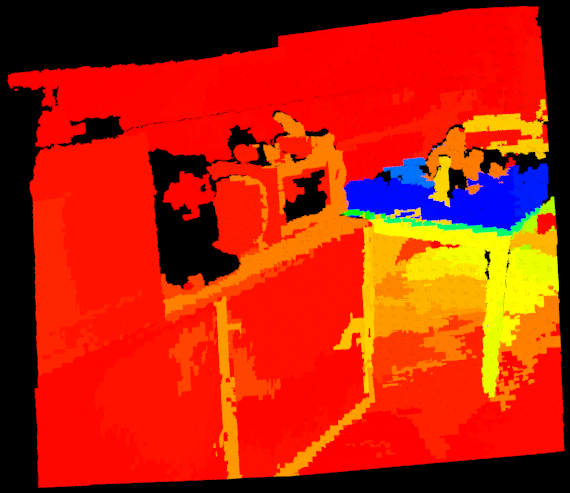} 
    \includegraphics[width=0.24\textwidth]{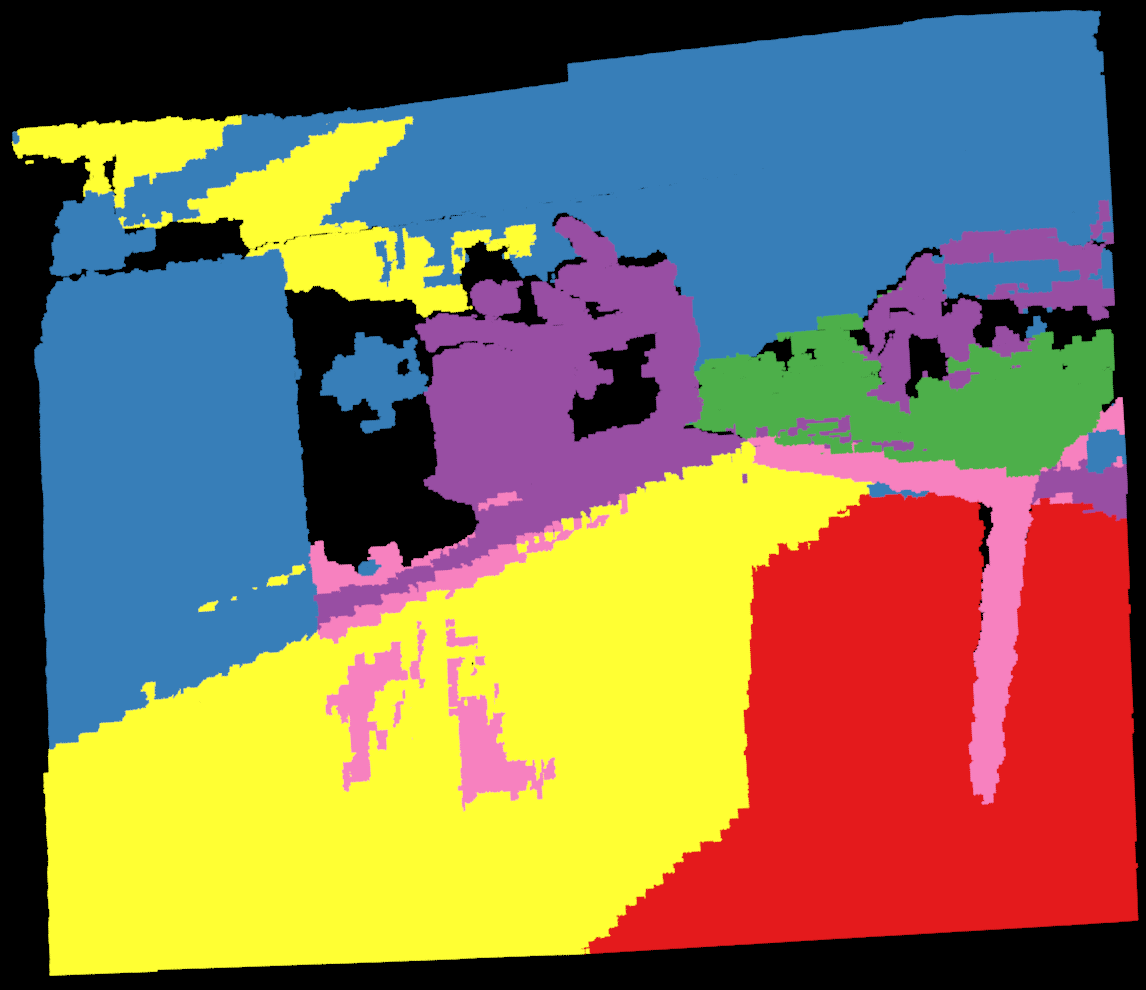} 
    \caption{Intermediate steps of our segmentation pipeline. Left to right: Input image, oversegmentation, conditional label probabilities (here for label \textit{table}, red=0, blue=1), final result after MRF. Color code given in Sec.~\ref{sec:results}.}
  \label{fig:pipeline}
\end{figure}

Especially since the emergence of cheap 3D sensor technology such as the Microsoft Kinect, semantic segmentation for indoor scenes has become a very active topic in the field of computer vision, and many different proposed methods show promising results.
However, using them in an interdisciplinary scope of computer vision and mobile robotics is very challenging due to the strict limitations imposed by mobile robotic systems. Many published scene segmentation algorithms are too complex to be executed on a real-time system, consequently there are not many applications actually making use of the results these methods provide yet.
Focused on this issue, in this paper we present an efficient and fast semantic segmentation framework developed and optimized to be deployed on a mobile service robot autonomously navigating in user apartments. As an example application, we show that our framework can be used to speed up the object search task previously described by inferring likely object locations from the segmentation results.

The remainder of this paper is structured as follows: In the next section we discuss recent developments in the area of semantic segmentation in computer vision and robotics. The proposed framework is presented in Sec.~\ref{sec:pipeline}, the used datasets to train and evaluate it are covered in Sec.~\ref{sec:datasets}. Sec.~\ref{sec:objectsearch} briefly introduces our mobile robot and describes the object search scenario in more detail, the results are discussed in Sec.~\ref{sec:results}. We finally give an overview of possible ideas for future work and conclude in Sec.~\ref{sec:conclusions}.
\vspace{-1.7mm}
\section{Related Work}
\vspace{-1.7mm}

The majority of proposed semantic segmentation algorithms \cite{Munoz2009a,Silberman2011,Anand2012,Valentin2013} is based on a similar architecture. In a first step, a clustering algorithm calculates an oversegmentation of the input scene and a feature vector is extracted for each cluster. The clusters are then classified according to the feature vector and in the last step a Conditional Random Field (CRF) or Markov Random Field (MRF) incorporates more global information to obtain the final labeling.

Munoz et al.\ \cite{Munoz2009a} proposed an outdoor scene labeling approach to label 3D points collected by a laser scanner. They learn the parameters of the CRF using a functional gradient algorithm.
To label sequences of frames, Floros et al. \cite{Floros2012} came up with a large CRF formulation connecting several subsequent frames to enforce time consistency in the labeling through higher-order potentials. As well as \cite{Munoz2009a}, their method is intended for outdoor scenarios, using a stereo camera setup.




With the arrival of new structured light depth sensors like the Microsoft Kinect, the attention shifted more towards the labeling of indoor scenes.
Silberman et al.\ \cite{Silberman2011,Silberman2012} presented the publicly available NYU Depth datasets, providing thousands of recorded RGB-D frames of different indoor scenes recorded with a Microsoft Kinect, many of them with densely annotated labels. Their baseline algorithm is based on 2D data taking into account depth and uses a neural network, followed by a CRF.
The first big improvement on the results of \cite{Silberman2011} was presented by Ren et al.\ \cite{Fox2012}, using kernel descriptors to describe patches around every pixel. They achieved the best results combining a segmentation tree and a superpixel MRF.
An alternative way of oversegmenting the input scene is used by Valentin et al.\ \cite{Valentin2013}. They calculate a mesh representation of the scene and compute feature vectors for all faces of the mesh, using geometric and color information.
Like \cite{Fox2012}, the approach presented by Couprie et al.\ \cite{Couprie2013a} omits the procedure of handcrafting suitable features to classify scene segments. They  exhaustively train a multiscale convolutional network to learn discriminative features directly from training data.
After downscaling regular Kinect frames by a factor of $2$, they are able to process more than $1$ frame per second.

Anand et al.\ \cite{Anand2012} formulate a large and expressive graphical model as a Mixed Integer Problem (MIP), encoding rich contextual information between scene segments.
Their method shows impressive results, but takes 2 minutes to find a solution for a single point cloud. A relaxed formulation of their problem can be solved much faster, however, they do not mention the total runtime including oversegmentation and feature extraction.
To our knowledge, their work is the first showing a direct application of semantic segmentation in the context of mobile robotics.

Recently introduced Decision Tree Fields (DTF) and Regression Tree Fields (RTF) are used by K\"ahler et al.\ \cite{Kahler2013} to classify scene segments. Especially RTFs achieved an appealing performance, as they offer fast inference times in addition to state-of-the-art results.

Unlike \cite{Munoz2009a,Floros2012,Silberman2011,Fox2012,Anand2012}, where the total processing time is not mentioned, the approaches presented in \cite{Valentin2013,Couprie2013a,Kahler2013} show fast inference times, what is an important aspect for robotics applications. However, with the exception of \cite{Couprie2013a}, they all rely on an input scene which has been densely reconstructed from several frames, which is contradicting the idea of a fast real-time system.
On the contrary, our proposed method works on a frame-by-frame basis without the need for a separate preprocessing stage.
\vspace{-1.7mm}

\section{Semantic Segmentation Pipeline}
\label{sec:pipeline}
\vspace{-1.7mm}

Our proposed point cloud processing pipeline consists of four steps, depicted in Fig.~\ref{fig:pipeline}. First, we create an oversegmentation of the scene, clustering it into many small homogeneous patches.
In the second step, we compute a manifold but efficient-to-compute feature set for each patch. 
The resulting feature vector is then processed by a classifier, which yields a probability for each patch being assigned a specific label. To that end, we use a Randomized Decision Forest (RDF), a classifier which is intensively discussed in \cite{Criminisi2013}.
In the last stage of our processing pipeline the classification results set up a pairwise Markov Random Field (MRF), whose optimization yields the final labeling. This last step smoothes the labeling out to correct ambiguous classification results due to noisy local patch information.
The final labeling then corresponds to the Maximum-a-posteriori of the output of the MRF.
\vspace{-1.7mm}
\subsection{Oversegmentation}
\label{sec:supervoxels}
\vspace{-1.7mm}

Like the majority of scene segmentation approaches, we first create an oversegmentation of the input data, such that the features can capture more information and classification is more robust against noise. Furthermore, this step drastically reduces the number of nodes for the final MRF stage, which results in much shorter inference times. To perform the segmentation, we use the supervoxel clustering algorithm proposed by Papon et al.\ \cite{Papon2013}.
%
%
\vspace{-1.7mm}
\subsection{Feature Calculation}
\label{sec:features}
\vspace{-1.7mm}

After the segmentation patches have been obtained, we calculate a set of features for each patch. Patches containing very few points are disregarded.
First, we calculate the three eigenvalues $\lambda_0 \le \lambda_1 \le \lambda_2$ of the scatter matrix of the patch. Then, similar to \cite{Munoz2009a}, we define three spectral features, namely \textit{pointness} ($\lambda_0$), \textit{surfaceness} ($\lambda_1-\lambda_0$) and \textit{linearness} ($\lambda_2-\lambda_1$). One of the most discriminative features is the height of the centroid of the patch above the ground plane. Additionally, we use the height values of the lowest and the highest point of the patch as a feature. Important information can also be extracted from the surface normals of a patch. Since they are already computed for the supervoxel clustering, we can add the angle of the mean surface normal of the patch with the ground plane and its circular standard deviation as features without increasing the computational complexity of the feature calculation stage. Finally, we also make use of the color information. We first transform all color values to the CIELAB color space and then store the mean color values of the patch and its respective standard deviations as the last two features. In total, we end up with a 14-dimensional feature vector $x$, which is then fed into the classification stage described in the following section.

\vspace{-1.7mm}
\subsection{Randomized Decision Forest}
\label{sec:randomforest}
\vspace{-1.7mm}

In recent years, Randomized Decision Forests and several variations of them \cite{Criminisi2013} have been successfully used for many different tasks in image processing and computer vision \cite{Schroff2008,Shotton2013full,Kahler2013}. They are capable of handling a large variety of different features, have a probabilistic output and are very efficient at training and at test time.

To train our RDF we follow the standard approach presented in \cite{Criminisi2013}, such that we end up with a pre-defined number of trees recursively splitting up the data with respect to the evaluation of randomly chosen split functions. Leaf nodes are created at the defined final depth level of the trees or if data cannot be split up any further. These nodes store the distribution of the labels of the training data which has reached the respective node.
At test time, a data point $x$ (i.e.\ feature vector) traverses all trees according to the learned split functions, starting at the root nodes, until it reaches a leaf node in every tree. The conditional probability $p(y|x)$ of label $y$ being assigned to a patch with feature vector $x$ is then defined as the mean of all label distributions stored in the reached leaf nodes.
\vspace{-1.7mm}
\subsection{Markov Random Field}
\label{sec:mrf}
\vspace{-1.7mm}
In the last stage of our processing pipeline we model the labeling problem with a Markov Random Field, similar to the formulation presented in \cite{Tombari2011}. An MRF is a graph-based model, where an \textit{undirected graph} is defined as a set $(\mathcal{V}, \mathcal{E})$, $\mathcal{V}$ denoting a set of vertices or nodes and $\mathcal{E}$ denoting a set of edges connecting nodes. In our case, each node $i \in \mathcal{V}$ corresponds to a patch and is assigned a label $y_i \in \mathcal{L}$, where $\mathcal{L}$ is the discrete set of label categories. The set of all label assignments is defined as $\mathcal{Y}=\{y_i\}$. We use a pairwise MRF, which means that we only consider edges connecting exactly two nodes. This allows us to directly infer $\mathcal{E}$ from the pairwise adjacency graph obtained by the supervoxel clustering, defining the set of nodes $\mathcal{A}_i$ connected to a node $i \in \mathcal{V}$. Following the Hammersley-Clifford theorem, the posterior probability of a label assignment $\mathcal{Y}$ is a Gibbs distribution, which can be reformulated as an energy function:
\begin{equation}
 E\left(\mathcal{Y}\right) = \sum_{i \in \mathcal{V}} \phi_i \left(y_i\right) + \sum_{i \in \mathcal{V}} \sum_{j \in \mathcal{A}_i} \phi_{i,j} \left(y_i,y_j\right)
 \label{eq:loggibbs}
\end{equation}
where $\phi_i \left(y_i\right)$ is the \textit{unary} term corresponding to the likelihood label $y_i$ being assigned to node $i$ and $\phi_{i,j} \left(y_i,y_j\right)$ is the \textit{pairwise} term corresponding to the pairwise likelihood of labels $y_i$ and $y_j$ being assigned to the nodes $i$ and $j$. The optimal labeling $\mathcal{Y}^*$ can be obtained by minimizing the energy function:
\begin{equation}
 \mathcal{Y}^* = \argmin_\mathcal{Y} E(\mathcal{Y})
\end{equation}
The unary term can directly be inferred from the conditional probabilities \pagebreak
calculated by the classification stage by transferring them to a cost:
\begin{equation}
 \phi_i \left(y_i\right) = \lambda \left(1 - p\left(y_i | x_i \right) \right)
 \label{eq:unary}
\end{equation}
where $\lambda$ is a weighting term defining the importance of the unary term compared to the pairwise term. For the pairwise term we use the common definition of the Potts model:
\begin{equation}
 \phi_{i,j} \left(y_i, y_j\right) =
 \begin{cases}
   0 & y_i = y_j \\
   e^{-\frac{\|p_i-p_j\|}{\sigma}}       & \text{otherwise}
  \end{cases}  
 \label{eq:pairwise}
\end{equation}
where $p_i$ and $p_j$ are the 3D coordinates of the centroids of the patches corresponding to the nodes $i$ and $j$. $\sigma$ regularizes the penalty assigned to an inconsistent labeling of adjacent patches. As proposed in \cite{Tombari2011}, we approximately solve the resulting optimization problem (\ref{eq:loggibbs}) using Loopy Belief Propagation \cite{Kim1983}.
\vspace{-1.7mm}
\section{Datasets for Training and Evaluation}
\label{sec:datasets}
\vspace{-1.7mm}
We train and evaluate our classifier on the NYU Depth V2 dataset published by Silberman et al.\ \cite{Silberman2012}. It contains densely labeled indoor scenes recorded with a Microsoft Kinect. In particular, a collection of $1{,}449$ frames has been labeled with more than $1{,}000$ classes. As the dataset has been recorded manually holding the camera, we need to fit a plane to all ``floor''-labeled points to retrieve the camera height and angles. If there are not enough floor points available in the image, it is discarded in the training and evaluation procedures.
For our purposes, we decided the most important object classes are larger structures commonly seen in apartments, as well as a separate object class. Therefore, we narrow down the labels available in the dataset to the set \textit{floor}, \textit{wall}, \textit{ceiling}, \textit{table}, \textit{chair}, \textit{cabinet}, \textit{object}, and \textit{unknown}.

Besides evaluating our framework on the popular NYU dataset, we also measure its perfomance on our own small dataset, consisting of 10 typical office scenes. The difference to the NYU dataset is that the point clouds have all been recorded from the same height and angle, a similar setting as on our mobile robot. The label set is the same as for the NYU dataset.
Some example images of the used datasets and the corresponding results are shown in Sec.~\ref{sec:results}.

\vspace{-1.7mm}
\section{Semantic Segmentation for Object Search}
\label{sec:objectsearch}
\begin{figure}[htbp]
  \centering
    \includegraphics[height=2.8cm]{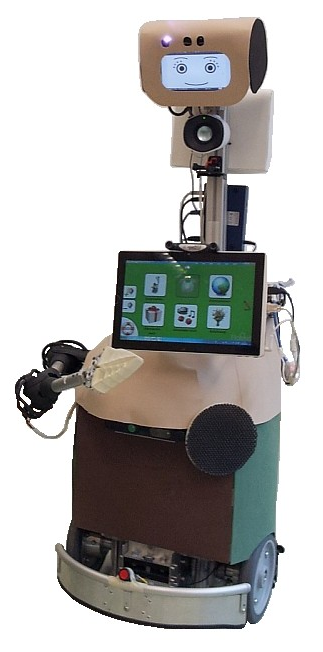}    
  \hspace{2cm}
    \includegraphics[height=2.8cm]{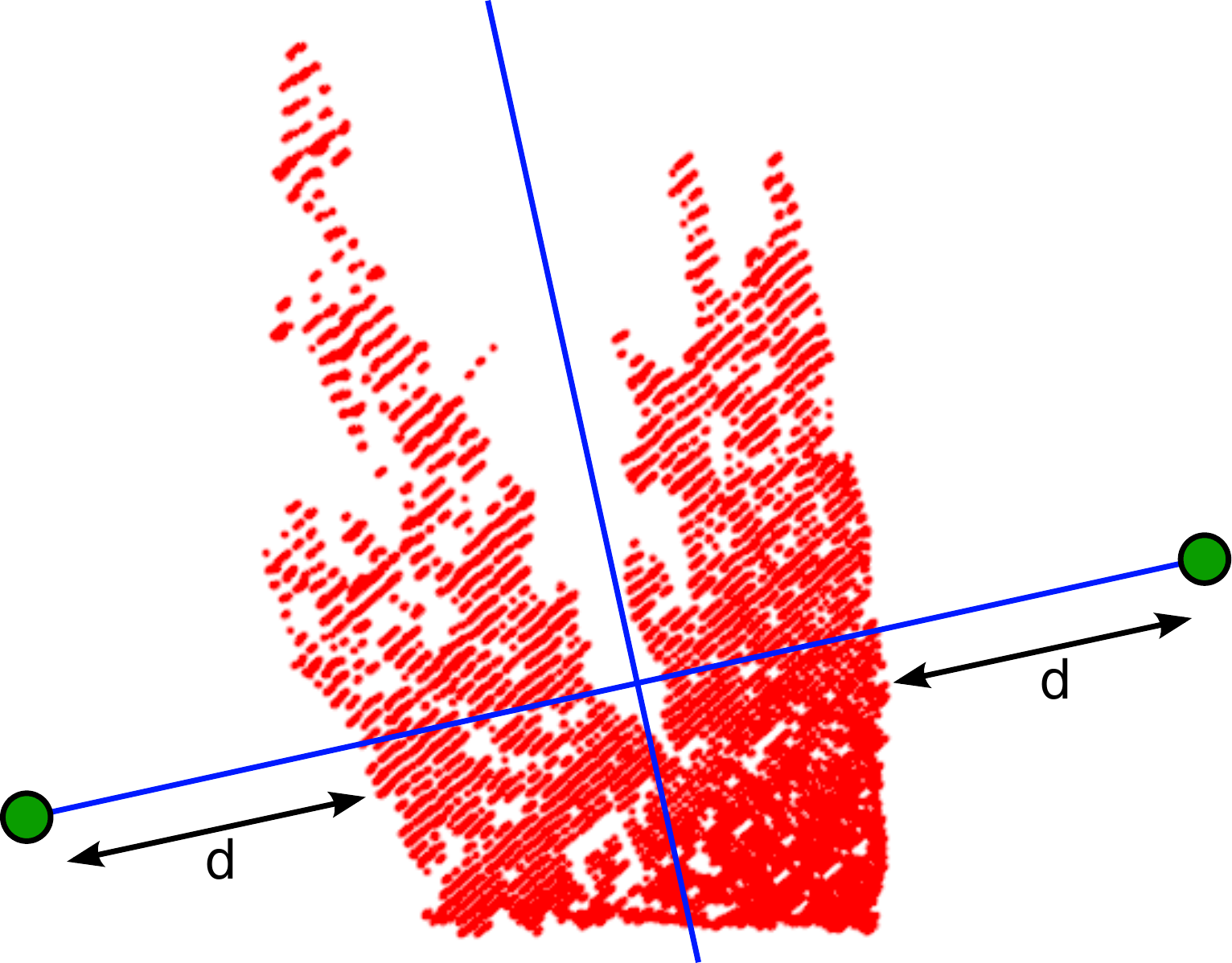}
  \caption{
  Left: The Hobbit robot. 
  Right: Heuristic to define search positions for a table cluster projected on the groundplane (red points). Blue lines: Principal axes of the cluster. The search positions (green dots) are placed on the second principal axis, adhering to a security distance $d$ from the table edge.}
  \label{fig:hobbitandsearchpositions}
\end{figure}

\vspace{-1.7mm}
In this section, we describe how we use our framework to speed up an object search task on the mobile service robot Hobbit \cite{Fischinger2013}. The robot is equipped with a differential drive and two RGB-D cameras. For our experiments only the camera in the head, which is mounted on a pan-tilt unit, is used. For manipulation the robot has an IGUS Robolink arm with a 2-finger gripping system. A picture of the platform can be seen in Fig.~\ref{fig:hobbitandsearchpositions} (left).

In the object search scenario the user asks the robot to search and fetch an object, e.g.\ a mug. The robot then sequentially navigates to a number of ``search positions'' defined in the map, where an object recognition algorithm is then run. So far, the list of search positions had to be pre-defined by an expert and stayed fixed. Using our framework this is no longer necessary, because likely object locations can directly be inferred from our segmentation results. In particular, because our robot can only grasp objects located on tables, we only consider positions next to large clusters of points labeled ``table'' as suitable object search positions. Consequently, after calculating the semantic labels of the current scene, we first use simple Euclidean clustering to obtain all tables in the scene. The resulting search positions are then defined by a simple heuristic, which is explained in Fig.~\ref{fig:hobbitandsearchpositions} (right). 
For further details about the whole object search scenario we refer to \cite{bajones2014}.
\vspace{-1.7mm}
\section{Results and Discussion}
\label{sec:results}
\vspace{-1.7mm}
To measure the labeling performance of our framework, we evaluated it with respect to the common multi-label metrics \textit{class average accuracy} and \textit{global accuracy}. The former is the mean of the diagonal of the confusion matrix, the latter is the mean of the pointwise accuracy over the whole test set. For the NYU dataset, we performed 5-fold cross-validation 
, for our own dataset we evaluated on all point clouds after training on the whole NYU dataset.
With a class average accuracy of $71.7\%$ and a global accuracy of $77.2\%$ for the NYU, respectively $55.6\%$ (class average) and $72.0\%$ (global) for our own dataset, our framework achieves superior performance with respect to the current state of the art \cite{Couprie2013a}. Of course, one has to keep in mind that we use a limited label set specific to our application, compared to other approaches evaluated the NYU dataset. An overview of all quantitative results is given in Table~\ref{tbl:results}, some qualitative examples are shown in Fig.~\ref{fig:nyuresults}.

We also evaluated how different parameters, namely the number of trees used in the RDF and the maximum tree depth, influence the accuracy. 
Fig.~\ref{fig:diagrams}~(left) shows that the accuracy significantly increases as soon as multiple trees are used in the RDF, but saturates if more than $8$ trees are used.
A similar effect can be observed for the maximum tree depth parameter, plotted in Fig.~\ref{fig:diagrams}~(right). With increased depth level, the RDF can capture the data structure better. However, due to the limited amount of training data, trees often do not grow deeper than $10$ levels in the training stage, which explains why the results do not improve for larger depth values.

Setting the maximum tree depth and the number of trees to $8$, our framework processes point clouds containing 640x480 points at a frame rate of about $1$\,fps on a 2.4\,GHz 8-core Intel i7 laptop. Regarding the operation on a robot, we consider this processing time to be sufficiently fast for many potential applications, such as the object search scenario described in Sec.~\ref{sec:objectsearch}.
\begin{table}[htb]
\caption{Class average and global accuracy of our framework in $\%$. We do not compare to the overall accuracies of \cite{Couprie2013a} because of the different label sets.}
\label{tbl:results}
 \begin{tabulary}{\textwidth}{@{}lCCCCCCCCCC@{}} 
 \toprule
   Dataset & &
\rot{Floor} &
\rot{Wall} &
\rot{Ceiling} &
\rot{Table} &
\rot{Chair} &
\rot{Cabinet} &
\rot{Object} &
   \rot{Avg.} & \rot{Global} \\
   \midrule
   \multirow{2}{*}{NYU V2} & \cite{Couprie2013a} & 87.3 & \textbf{86.1}& 62.6 & 10.2 &34.1 &- &8.7 &- &- \\
                           & Ours & \textbf{96.9} & 75.0 & \textbf{92.3}&\textbf{59.7} &\textbf{72.2} &\textbf{58.4} &\textbf{40.0} &\textbf{71.7} &\textbf{77.2}\\
   \midrule
   Our data  & Ours & \textbf{98.0}&\textbf{87.8} & - & \textbf{92.2}& \textbf{62.8}& \textbf{31.5}& \textbf{16.4}& \textbf{55.6}& \textbf{72.0} \\  
   \bottomrule
 \end{tabulary} 
\end{table} 
\begin{figure}[htbp]
  \centering
    \includegraphics[width=0.48\textwidth]{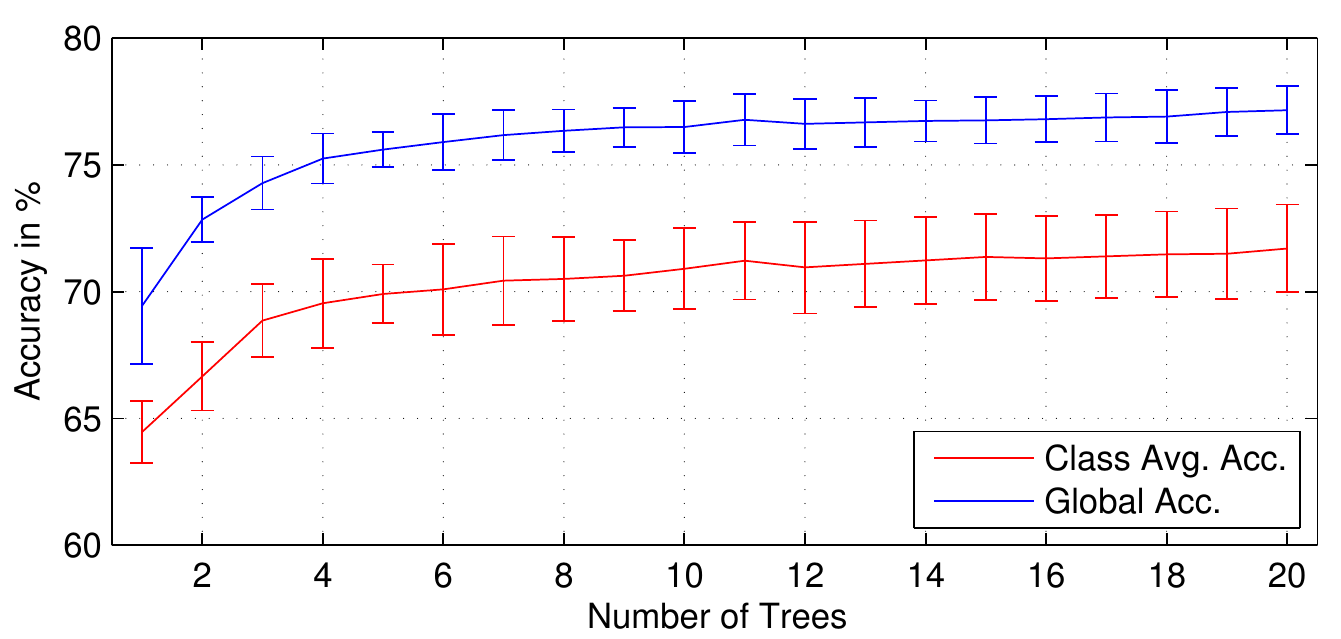}              
  \quad  
     \includegraphics[width=0.48\textwidth]{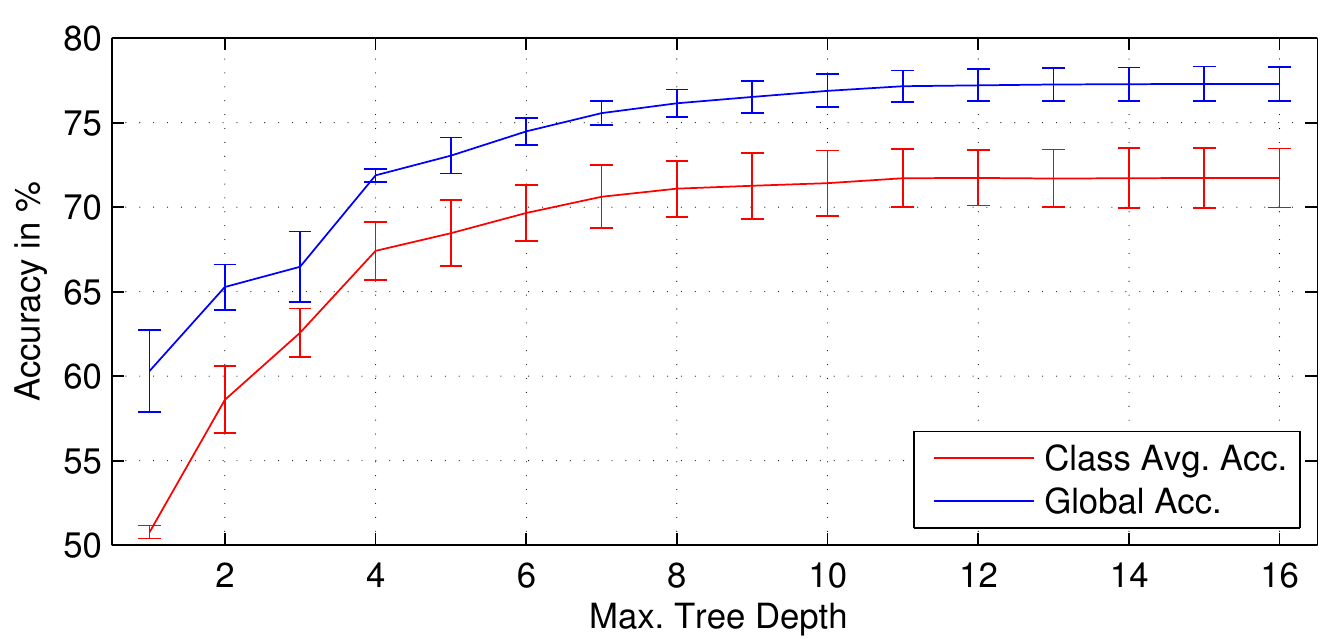}    
  \caption{Influence of RDF parameters \textit{number of trees} and \textit{max tree depth} on the labeling accuracies for the NYU Depth V2 dataset.}
  \label{fig:diagrams}
\end{figure}
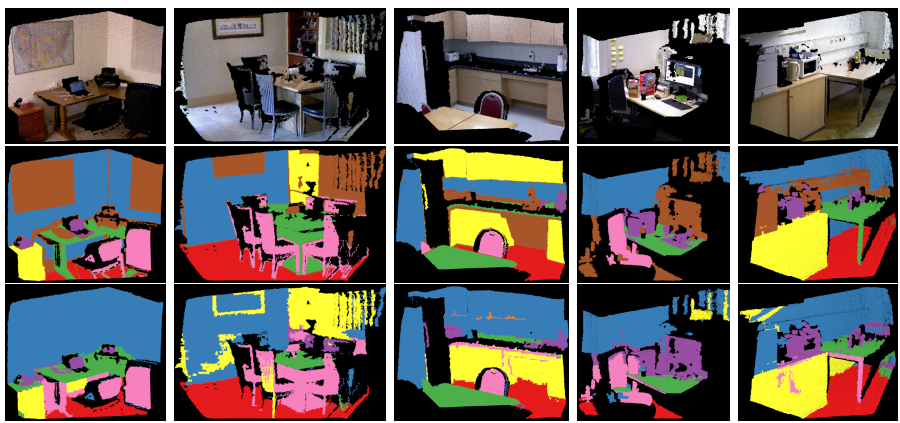
\begin{figure}[htb]
  \centering
  \includegraphics[height=1.8cm]{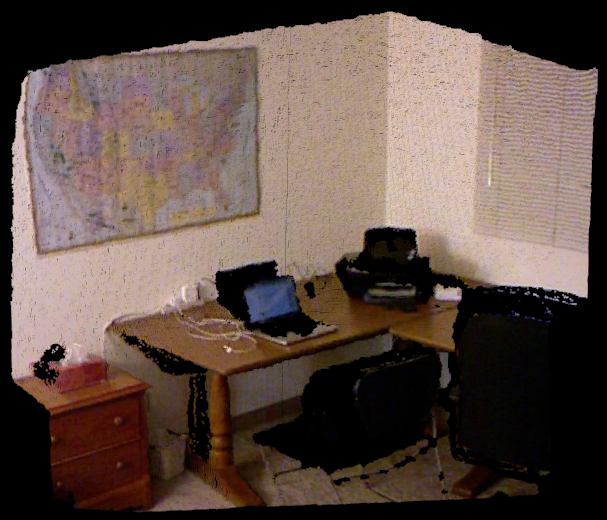}
  \includegraphics[height=1.8cm]{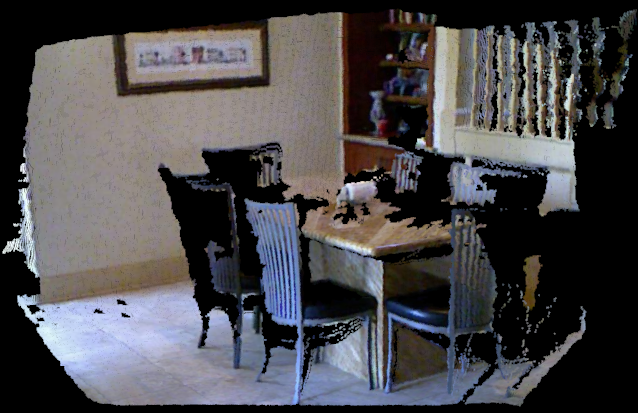}
  \includegraphics[height=1.8cm]{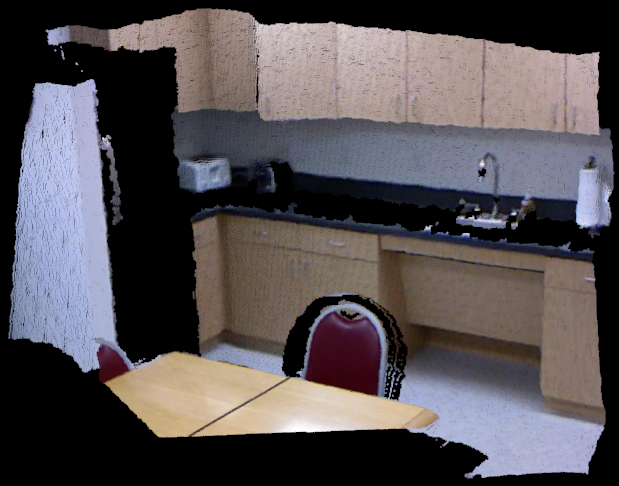}
  \includegraphics[height=1.8cm]{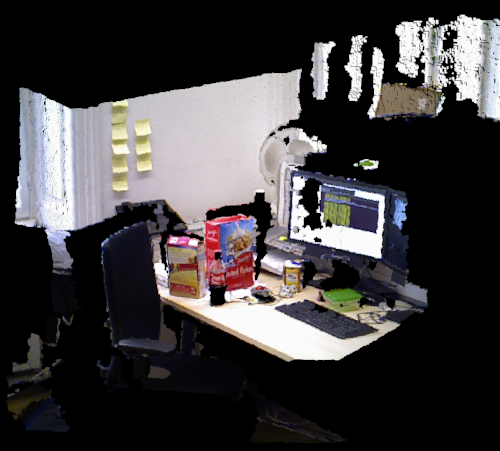}
  \includegraphics[height=1.8cm]{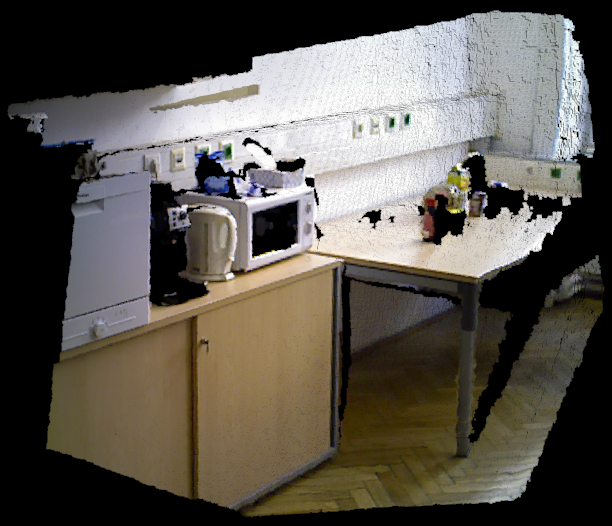} \\  
  \includegraphics[height=1.8cm]{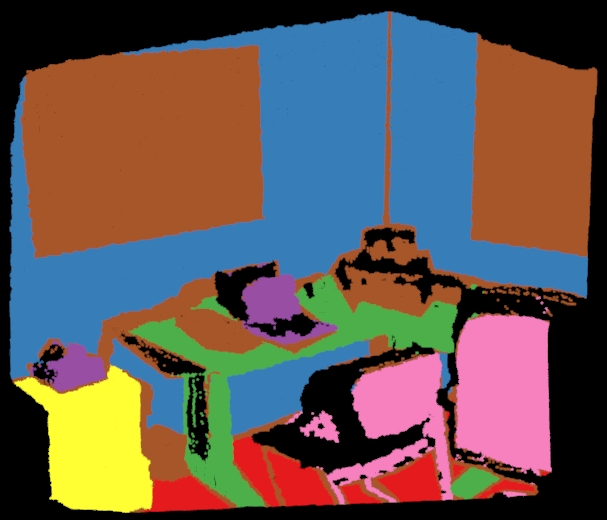}
  \includegraphics[height=1.8cm]{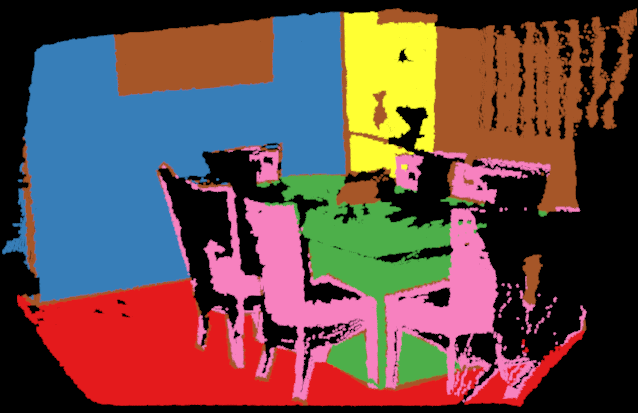}
  \includegraphics[height=1.8cm]{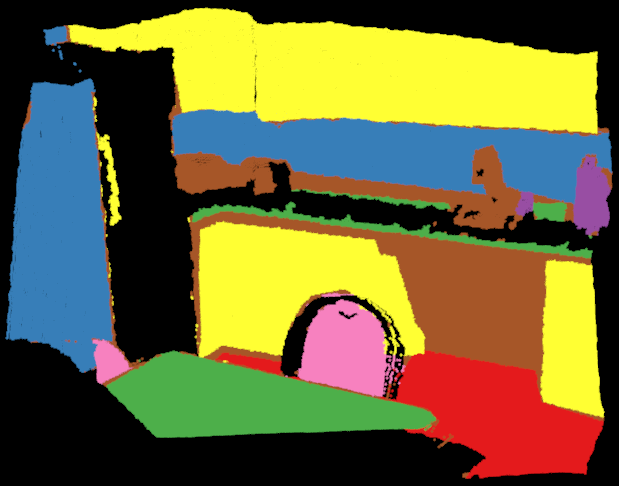}
  \includegraphics[height=1.8cm]{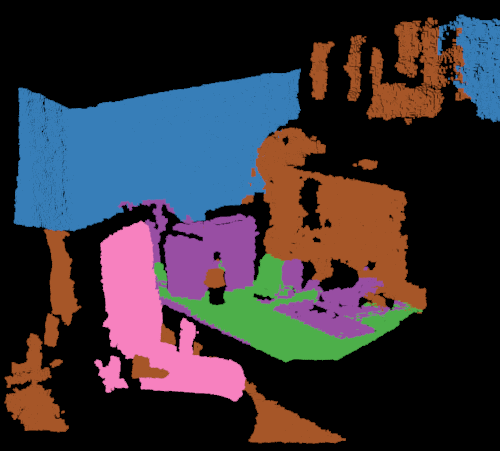}
  \includegraphics[height=1.8cm]{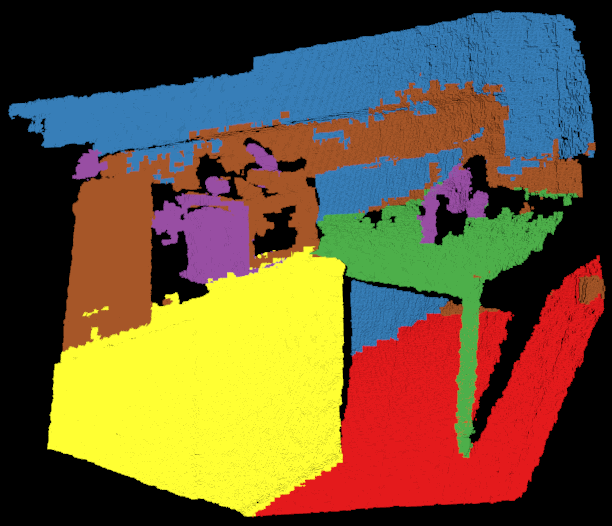} \\
  \includegraphics[height=1.8cm]{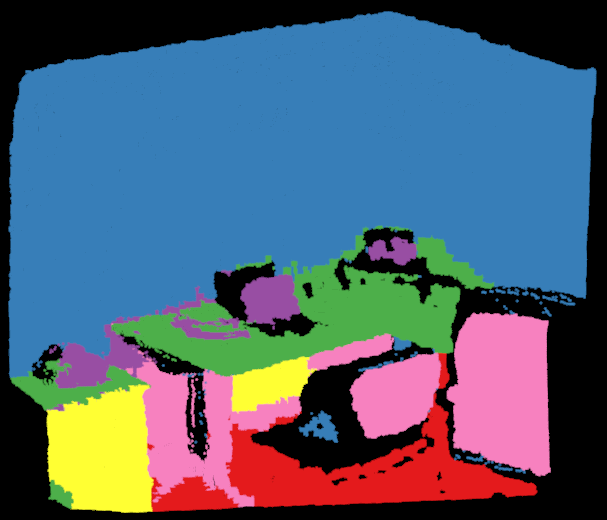}
  \includegraphics[height=1.8cm]{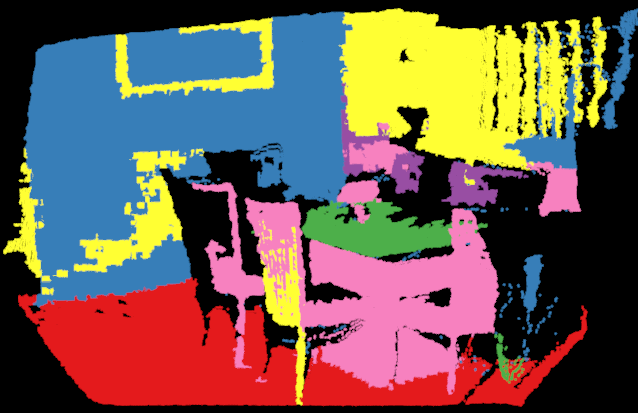}
  \includegraphics[height=1.8cm]{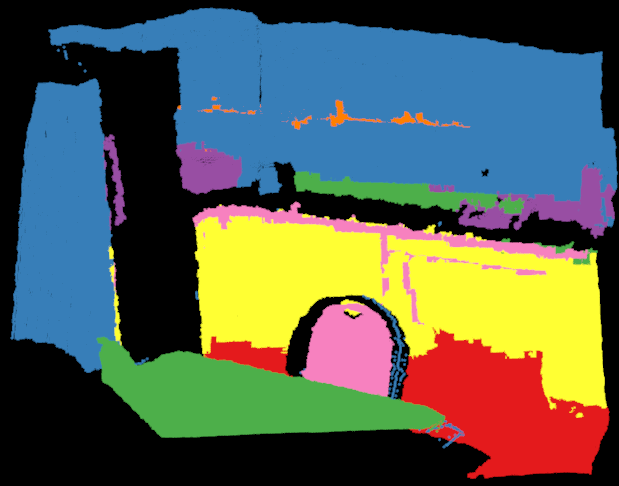}
  \includegraphics[height=1.8cm]{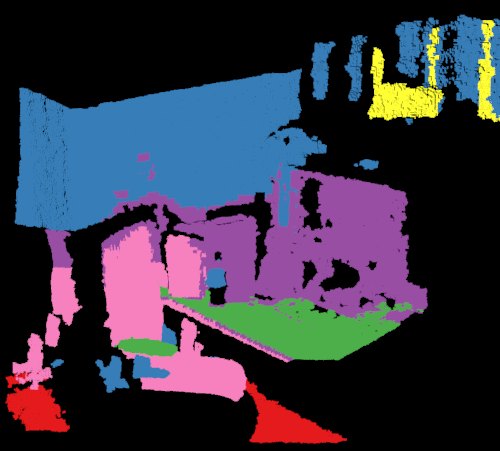}
  \includegraphics[height=1.8cm]{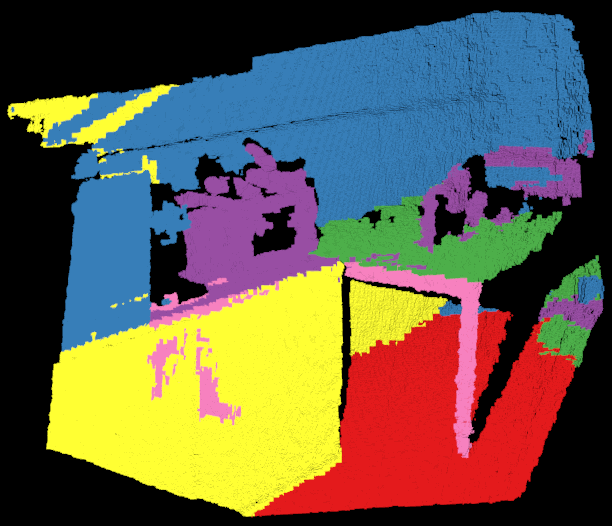} \\
  \begin{tikzpicture}\filldraw[draw=black,fill=floor] (1,1) rectangle (1.4,1.25); \end{tikzpicture} floor
  \begin{tikzpicture}\filldraw[draw=black,fill=wall] (1,1) rectangle (1.4,1.25); \end{tikzpicture} wall 
  \begin{tikzpicture}\filldraw[draw=black,fill=ceiling] (1,1) rectangle (1.4,1.25); \end{tikzpicture} ceiling
  \begin{tikzpicture}\filldraw[draw=black,fill=table] (1,1) rectangle (1.4,1.25); \end{tikzpicture} table
  \begin{tikzpicture}\filldraw[draw=black,fill=chair] (1,1) rectangle (1.4,1.25); \end{tikzpicture} chair
  \begin{tikzpicture}\filldraw[draw=black,fill=cabinet] (1,1) rectangle (1.4,1.25); \end{tikzpicture} cabinet
  \begin{tikzpicture}\filldraw[draw=black,fill=object] (1,1) rectangle (1.4,1.25); \end{tikzpicture} object
  \begin{tikzpicture}\filldraw[draw=black,fill=unknown] (1,1) rectangle (1.4,1.25); \end{tikzpicture} unknown
  \caption{Example results for NYU Depth V2 (first 3 columns) and our (last 2 columns) dataset. Top: Input image. Middle: Groundtruth. Bottom: Results.}
  \label{fig:nyuresults}   
\end{figure}
\vspace{-1.7mm}
\section{Conclusions and Outlook}
\label{sec:conclusions}
\vspace{-1.7mm}
We presented an efficient 3D semantic segmentation framework for indoor scenes. We showed that our method, based on an RDF, achieves accurate results at a frame rate feasible for the application on a mobile robot. We demonstrated that by successfully deploying our framework on a mobile service robot, where we used our method to detect possible object locations in a room and in turn are able to dynamically infer a more efficient object search procedure.

Still, there are aspects of our framework which could potentially be improved. So far, we do not make use of the contextual relationship between scene segments. By incorporating simple pairwise features, e.g.\ color and height difference, we expect the accuracy to further increase, especially for similar labels such as wall and cabinet. We also plan to investigate further exploitations of semantic segmentation in the scope of mobile robotics. An interesting application could be the construction of a complete semantic map, fusing labeling results from different viewpoints. This map would not only encode more information, but should also be more robust against noisy classification results.

\vspace{-1.7mm}
\section*{Acknowledgments}
\vspace{-1.7mm}
This work has been partially funded by the European Commission under grant agreements FP7-IST-288146 HOBBIT, FP7-IST-600623 STRANDS and the Austrian Science Foundation (FWF), Project I513-N23 vision@home.
\vspace{-1.7mm}
\bibliography{oagm}
\end{document}